\pdfoutput=1
\documentclass[11pt,a4paper]{article}
\usepackage[hyperref]{naaclhlt2018}
\usepackage{times}
\usepackage{latexsym}
\usepackage{algorithm}
\usepackage{algorithmic}
\usepackage{amsmath,amssymb}
\usepackage{graphicx}
\usepackage{multirow}

\usepackage{url}

\aclfinalcopy 
\def\aclpaperid{895} 

\DeclareMathOperator*{\argmin}{argmin}

\title{Understanding and Improving Multi-Sense Word Embeddings \\ via Extended Robust Principal Component Analysis}

\author{
Haoyue Shi $^{\dagger}$,
Yuqi Sun $^{\dagger}$, 
Junfeng Hu $^{\S, \ddagger}$\\
  $\ddagger$ School of Electronics Engineering and Computer Science, Peking University \\
  $\S$ Key Laboratory of Computational Linguistics (Ministry of Education), Peking University\\
  {\tt \{hyshi,sun\_yq,hujf\}@pku.edu.cn}\\}

\date{}

\begin{document}
\maketitle
\begin{abstract}
Unsupervised learned representations of polysemous words generate a large of pseudo multi senses since unsupervised methods are overly sensitive to contextual variations. In this paper, we address the pseudo multi-sense detection for word embeddings by dimensionality reduction of sense pairs. We propose a novel principal analysis method, termed Ex-RPCA, designed to detect both pseudo multi senses and real multi senses. With Ex-RPCA, we empirically show that pseudo multi senses are generated systematically in unsupervised method. Moreover, the multi-sense word embeddings can by improved by a simple linear transformation based on Ex-RPCA. Our improved word embedding outperform the original one by 5.6 points on Stanford contextual word similarity (SCWS) dataset. We hope our simple yet effective approach will help the linguistic analysis of multi-sense word embeddings in the future.
\end{abstract}

\section{Introduction}
Multi-sense word embedding is widely applied for learning representations of polysemous words.
In general, existing works can be divided into two groups: through \textit{unsupervised learning}  and \textit{semi-supervised learning}.

Most of unsupervised methods exploit contextual information as the guidance for sense clustering~\cite{reisinger2010multi,huang2012improving,neelakantan2015efficient,guo2014learning,li2015multi,iacobacci2015sensembed,cheng2015syntax,lee2017muse}.
As shown in~\newcite{shi2016real}, nevertheless, these methods are so sensitive to contextual variations that they often embed one single sense into several vectors, also referred to as \textit{pseudo multi-senses}.
As for supervised methods~\cite{chen2014unified,cao2017bridge}, external knowledge base, \textit{e.g.}, WordNet~\cite{miller1995wordnet} is utilized to define senses. 
The limitation of such supervised methods is that it requires manually designed senses of words.

In this paper, we aim to address the pseudo multi-sense problem associated with \textit{unsupervised} methods.
Since pseudo multi-senses constitute a overwhelming majority of all sense pairs~\cite{shi2016real}, intuitively, if we construct a matrix of which columns are the differences of sense vectors of each word, dimensionality reduction algorithms such as principal component analysis (PCA)~\cite{hotelling1933analysis,jolliffe1986principal} can be utilized to decompose it into two additive terms, \textit{i.e.}, a low-rank matrix and a Gaussian noise term \cite{mika1999kernel}.
As we will show, the column space of the low-rank matrix represents the most salient ``pseudo multi-sense directions'', while the error term stands for the random error occurred during training of neural word embeddings. 

Unfortunately, PCA ignores the effect of real multi-sense, of which the vector representations should have salient difference~\cite{neelakantan2015efficient}, and their vector differences shall no longer be viewed as Gaussian noise of pseudo multi-sense directions.
Inspired by robust PCA (RPCA)~\cite{wright2009robust}, we propose an effective method, termed extended robust PCA (Ex-RPCA), which is able to fit the dense property of word embeddings by jointly considering a Gaussian noise (random noise) and a sparse yet large noise (salient difference noise). We also propose two solutions for the optimization of Ex-RPCA.

With the principal components of sense-wise difference matrix, a transformation matrix is generated based on the objective that the vector representations of pseudo multi-senses should be as closed as possible. 
We construct the transformation matrix with the property that its kernel space is spanned by principal components in PCA or Ex-RPCA. 
The final multi-sense word embeddings can be obtained by a simple linear transformation with the transformation matrix.  Our main contributions are as follows:

1. We frame the pseudo multi-sense detection for word embeddings as a dimensionality reduction problem. We empirically show the extracted principal components reflect the dominant directions of pseudo multi-sense pairs.

2. A new principal component analysis algorithm, namely Ex-RPCA, is proposed to address the issue of robustness of PCA. Ex-RPCA is able to tolerate large noise and also fit the property of word embeddings. Moreover, Ex-RPCA serves as a good indicator for linguistic analysis of multi-sense word embeddings.

3. We evaluate the proposed framework on SCWS dataset. Significantly, we improve performance on contextual word similarity task over a strong reference \cite{neelakantan2015efficient} by a margin of absolute 5.6 points.

\section{Problem Formulation}
For multi-sense word vector set $V$ with $n$ words, let $\mathbf{v}_s^w \in V$ denote the vector of the $s^{th}$ sense of word $w$, and let $n_w$ denote the sense number of $w$. 
Define sense-wise difference matrix $ M = (\mathbf{v}_1^{w_1} - \\ 
\mathbf{v}_2^{w_1}, \mathbf{v}_2^{w_1} - \mathbf{v}_1^{w_1}, \mathbf{v}_1^{w_1} - \mathbf{v}_3^{w_1}, \mathbf{v}_3^{w_1} - \mathbf{v}_1^{w_1}, \cdots , \\
\mathbf{v}_{n_{w_1}-1}^{w_1} - \mathbf{v}_{n_{w_1}}^{w_1}, \mathbf{v}_{n_{w_1}}^{w_1} - \mathbf{v}_{n_{w_1}-1}^{w_1}, \mathbf{v}_1^{w_2} -  \mathbf{v}_2^{w_2}, \mathbf{v}_2^{w_2} - \mathbf{v}_1^{w_2}, \cdots , \mathbf{v}_{n_{w_n}-1}^{w_{n}} - \mathbf{v}_{n_{w_n}}^{w_{n}},\mathbf{v}_{n_{w_n}}^{w_{n}} - \mathbf{v}_{n_{w_n}-1}^{w_{n}}) $
, which has $\sum_{w \in V}n_{w}\times({n_{w}}-1)$ columns. 
PCA with respect to the columns of $M$ decomposes the matrix into two additive terms: 
\begin{equation}
M = L + E
\label{eq:PCA}
\end{equation}
where $L$ is a low-rank matrix and $E$ is the Gaussian noise term.

However, PCA is not robust enough to handle \textit{strong} noise.
Hence, robust PCA (RPCA) is introduced~\cite{wright2009robust} to decompose a matrix to a low-rank term and a sparse noise term:

\vspace{-0.4cm}
\begin{equation}
\setlength{\abovedisplayskip}{0.1cm}
\setlength{\belowdisplayskip}{-0.25cm}
M = L + S
\end{equation}
\vspace{-0.7cm}

The sparse term is able to tolerate extremely strong noise. RPCA can be solved via convex optimization. 
Unfortunately, RPCA is not able to reduce matrix to an explicitly fixed dimensionality, as what we can do with the matrix in PCA. 

Therefore, we propose the extend RPCA and will show an iterative solution to tackle this issue:
\begin{equation}
\begin{aligned}
\min_{L, E, S} & ~~ {\rm rank}(L) + \lambda_1 ||E||_F^2 + \lambda_2 ||S||_0 \\
{\rm s.t.} & ~~ M = L + E + S
\end{aligned}
\label{eq:formulation}
\end{equation}
where $\lambda_1$ and $\lambda_2$ are weights for the noise terms.
Let $V^e=\{\mathbf{v}^e_0, \mathbf{v}^e_1, \cdots, \mathbf{v}^e_k\}$ denote the set of the first $k$ principal components extracted by Ex-RPCA and let $V^p$ denote the set extracted by PCA.  
We will empirically show that vectors in both $V^p$ and $V^e$ represent dominant directions of pseudo multi-sense pairs.

\label{sec: 3.1}

\section{Our Approach}
In this section, we propose two solutions to Ex-RPCA problem from different angles of view, and describe the algorithm for elimination of pseudo multi-sense directions using linear transformation. 

\subsection{Convex Optimization}
Following \newcite{wright2009robust}, we re-formulate Eq. \eqref{eq:formulation} to a convex-optimization problem:
\begin{equation}
\setlength{\abovedisplayskip}{3pt}
\setlength{\belowdisplayskip}{3pt}
\begin{aligned}
\min_{L, E, S} & ~~ ||L||_* + \lambda_1 ||E||_F^2 + \lambda_2 ||S||_1 \\
{\rm s.t.} & ~~ M = L + E + S
\end{aligned}
\label{eq:convex-formulation}
\end{equation}
which can be solved by inexact augmented Lagrange method \cite{lin2011linearized} (Appendix \ref{append: solution}).

While solving Problem~\eqref{eq:convex-formulation}, we cannot directly control the ${\rm rank}(L)$ instead of controlling $\lambda_1$ and $\lambda_2$, either.
In order to seek a fair comparison against PCA which can set ${\rm rank}(L)$ as a hyper-parameter, we introduce an iterative solution. 

\subsection{Iterative Solution via PCA}
In Eq.~\eqref{eq:formulation},
$E$ is a zero-averaged $i.i.d.$ Gaussian noise term with variance $\sigma^2$. 
If we apply PCA to reduce the column space of $M$ to less than 5 dimensions, the percentage of elements in $E$ which are outside the range of $(-3\sigma_E, 3\sigma_E)$ is $2.7\%$, where $\sigma_E$ is the standard deviation of $E$.
This is much higher than the expected percentage ($0.3\%$) based on three-sigma rule of Gaussian distribution.

To overcome the issue, we propose an iterative solution for Ex-RPCA. The motivation is to erase the strong noise of $E$ gradually, with ${\rm rank}(L)$ fixed as a hyper-parameter.
At the $t^{th}$ iteration of the iterative solution, we first apply PCA to decompose $M^{(t)}$ into $L^{(t)}+E^{(t)}$, as described in Eq \eqref{eq:PCA}.
After that, we extract the sparse noise $S^{(t)}$ from $E^{(t)}$. 
We compute a 0-1 matrix $P^{(t)}$ to mask the large noise elements in $E^{(t)}$ by:
\begin{equation}
\setlength{\abovedisplayskip}{0pt}
\setlength{\belowdisplayskip}{3pt}
P^{(t)}_{i,j} = \left\{ \begin{aligned} 
&0,~~ -3\sigma_{E^{(t)}} \leq E^{(t)}_{i,j} \leq 3\sigma_{E^{(t)}} \\
&1,~~ otherwise
\end{aligned}
\right.
\label{eq: mask}
\end{equation}
and compute the sparse noise by 
\begin{equation}
\setlength{\abovedisplayskip}{3pt}
\setlength{\belowdisplayskip}{3pt}
S^{(t)} = P^{(t)} \circ E^{(t)}
\label{eq: sparse}
\end{equation}
where $\circ$ represents element-wise product of two matrices.
After obtaining the sparse noise term $S^{(t)}$, let $M^{(t+1)} = M^{(t)} - S^{(t)}$ and go to the next iteration.
Algorithm \ref{alg: iterative-ex-rpca} summarizes the inexact iterative solution for Ex-RPCA.
The proof of convergence of this algorithm is trivial.
\begin{algorithm}[!t]
\begin{algorithmic}[1]  
\REQUIRE{sense-wise difference matrix $M$, dimensionality of principal components $d$\\}
\ENSURE{low-rank matrix $L$, sparse noise matrix $S$, Gaussian noise matrix $E$\\}
\STATE{Let $M^{(0)}=M, t=0, S= $ zero matrix}
\WHILE{{\sl not converged}}
 \STATE{Compute $L^{(t)} + E^{(t)} = M^{(t)}$ using PCA}
 \STATE{Compute $P^{(t)}$ using Eq \eqref{eq: mask}}
 \STATE{Compute $S^{(t)}$ using Eq \eqref{eq: sparse}}
 \STATE{Let $S = S + S^{(t)}$}
 \STATE{Let $M^{(t+1)} = M^{(t)} - S^{(t)}$}
 \STATE{Let $t = t + 1$}
\ENDWHILE
\STATE{\textbf{return} $L^{(t)}, E^{(t)}, S$}
\end{algorithmic}
\caption{Iterative solution for Ex-RPCA.}
\label{alg: iterative-ex-rpca}
\end{algorithm}

\subsection{Pseudo Multi Sense Elimination}
\label{sec: linear-transformation}
With the extracted principal components, we can directly calculate a linear transformation to eliminate the pseudo multi senses of word embeddings. Based on the property of kernel space, there exists a unique linear transformation which satisfies that: 
(1) Each principal component is projected to $\mathbf{0}$.
(2) Each vector orthogonal to all principal components is projected to itself. 
Due to the limited  space, the proof is in Appendix \ref{append: pseudo-elimination}.

With the linear transformation, the original word embeddings $V$ can be projected to a new space

\begin{equation}
\setlength{\abovedisplayskip}{3pt}
\setlength{\belowdisplayskip}{3pt}
\tilde{V} = \{ \tilde{\mathbf{v}}_s^w = T \mathbf{v}_s^w  | \mathbf{v}_s^w \in V\}
\end{equation}

In the new space $\tilde{V}$, the effect of pseudo multi senses is much lower than that in the original space $V$.

\section{Experiments}
The experiment section is organized as follows: we first analyze the properties of the principal components extracted by PCA and Ex-RPCA in Section~\ref{subsec:principal}; then we qualitatively analyze the sparse noise term of Ex-RPCA in Section~\ref{subsec:sparse}; finally, we demonstrate the effectiveness of the proposed methods in Section~\ref{subsec:results} on contextual word similarity task.

\begin{table}[t]
\centering
\begin{tabular}{ll|p{0.35\textwidth}}
\hline
\textbf{Model} & \textbf{\#} & \textbf{Pairs} \\
\hline
\multirow{3}{2pt}{PCA} &1 & income$_{2,4/4,5}$, campaigns$_{1,5}$, age$_{6,7}$, development$_{4,5,/1,4/2,5}$, goals$_{2,6}$  \\
& 2&  Berlin$_{0,6}$, Martin$_{3,4}$, Greek$_{0,3}$, Jan$_{0,4/0,6}$, name$_{1,3}$ \\
& 3& quarterback$_{3,9}$,  playoff$_{3,9/1,9/6,9}$，  NBA$_{0,1}$, Houston$_{1,3}$, mayor$_{0,6}$\\
\hline
\multirow{3}{2pt}{Ex-RPCA, rank({\sl L}) =3} & 1&  after$_{1,2}$,  eventually$_{0,1}$, whilst$_{0,1}$, again$_{1,2}$, finally$_{1,2}$\\
& 2& although$_{1,2}$,  well$_{2,3}$, initially$_{1,2}$, more$_{1,4}$, both$_{0,2}$\\
& 3& Brian$_{1,2}$, February$_{0,2}$, Daniel$_{1,2}$, September$_{2,7}$, Frank$_{0,2}$\\
\hline
\end{tabular}
\caption{\label{table: sense-pair-projection} Representative sense pairs for principal components. The subscripts refer to sense ids in pairs.}
\end{table}
\begin{table}[t]
\centering
\begin{tabular}{lrrr}
\hline
\textbf{Component} & \textbf{Feature} & \textbf{Avg. cos} & \textbf{$\mathbf{\rho_{var}}$}\\
\hline
PCA\#1 & political words & 0.27 & 12.3\\
PCA\#2 & proper nouns & 0.22 & 8.9\\
PCA\#3 & sports related& 0.20 & 5.8\\
sum$_{\rm PCA}$ & N/A & 0.69 & 27.0 \\
\hline
Ex-RPCA\#1 & conj. or adv.& \textbf{0.76} & 13.7\\
Ex-RPCA\#2 & conj. or adv.& 0.72 & 9.9\\
Ex-RPCA\#3 & proper nouns & 0.64 & 6.4\\
sum$_{\rm Ex-RPCA}$ & N/A & \textbf{2.12} & \textbf{30.0}\\
\hline
\end{tabular}
\caption{\label{table:common-features} Manually summarized common features of senses listed in Table \ref{table: sense-pair-projection} for each principal component, average cosine value of shown senses to each principal component, and explained variance ratio $(\times 100)$ of each principal component.}
\end{table}
\subsection{Analysis of Principal Components}
\label{subsec:principal}
In Table~\ref{table: sense-pair-projection}, we show the words of which sense pairs have the top-5 largest cosine similarity with the first three principal components extracted by PCA and Ex-RPCA. 
Interestingly, most of the words listed do not have multiple senses, \textit{e.g.}, ``Berlin" for PCA and ``eventually" for Ex-RPCA, yet they are assigned with several sense ids. 
For those words with multiple senses, we manually investigate what sense it represents through its nearest neighbors.
With no exception, the listed sense pairs are all pseudo multi-senses. 
This phenomenon indicates that pseudo multi-senses are generated systematically rather than accidentally.

Table~\ref{table:common-features} collects the manually summarized common features of sense pairs related to each principal components.
According to the results, PCA is more sensitive to senses with specific meanings, while Ex-RPCA is able to capture part of speech (POS) features, which is considered more general than topical features captured by PCA. 
Experimental results demonstrate that Ex-RPCA is more capable to discover systematical relations.

\subsection{Analysis of Sparse Noise in Ex-RPCA}
\label{subsec:sparse}
Ex-RPCA decomposes a matrix into a low-rank term for pseudo multi-senses, a Gaussian error term for errors randomly occurred during training, and a sparse noise term for salient difference of real multi-senses. We choose several words and select two senses of each word, manually infer the meaning of each sense through nearest neighbors, and compare the 2-norm of the corresponding column vector in sparse noise matrix. As shown in Table~\ref{table: knn}, the pairs of real multi senses (\textit{i.e.}, two senses of word ``prime" and ``yard") have larger noise than pairs of pseudo multi senses (\textit{\textit{i.e.}}, two senses of word ``cat"). As for the word ``engine", we can infer that there is indeed a slight difference between its two senses, although people sometimes prefer combining them into a single sense. From our observation, the sparse noise term in the proposed Ex-RPCA can serve as an accurate indicator of real multi senses.

\begin{table}[t]
\centering
\begin{tabular}{p{0.07\textwidth}p{0.32\textwidth}p{0.02\textwidth}}
\hline
\textbf{Word} & \multicolumn{2}{l}{
\textbf{Nearest Neighbors}~~~~~~~~~~~~~~~~~~~~~ $\mathbf{||S_{p}||_2}$} \\ 
\hline
${\rm prime}_{s0}$ & minister, cabinet, parliament\\
${\rm prime}_{s1}$ & modulo, space, equivalently, real & 3.35\\
\hline
${\rm yard}_{s0}$ & touchdown, interception, kickoff\\
${\rm yard}_{s1}$ & lawn, backyard, garden, porch & 2.75\\
\hline
${\rm engine}_{s0}$ & jeep, truck, wheel, vehicle, car\\
${\rm engine}_{s1}$ & camshaft, turbine, gearbox & 0.61\\
\hline
${\rm cat}_{s0}$ & dog, pet, wolf, bird, animal \\
${\rm cat}_{s1}$ & dog, pets, puppy, cats, fox & ~~~~~~0\\
\hline
\end{tabular}
~\\~\\
\begin{tabular}{lll}
\hline
\textbf{Word} & \textbf{Sense \#0} & \textbf{Sense \#1}\\ 
\hline
prime & (political) role & a kind of number\\ 
yard & unit of measure & outdoor enclosure\\
engine & railroad locomotive & machine \\ 
cat & animal & animal \\
\hline
\end{tabular}

\caption{\label{table: knn} Above: k nearest neighbors of senses in random selected pairs, as well as the 2-norm of the corresponding columns in noise matrix. 
Below: inferred meaning for each sense.}
\end{table}

\subsection{Word Similarity}
\label{subsec:results}
To further demonstrate the effectiveness of our framework, we choose WordSim-353 (WS-353) dataset~\cite{finkelstein2001placing} and Stanford Contextual Word Similarity (SCWS) dataset ~\cite{huang2012improving} for quantitative evaluation. 

WS-353 dataset provides a list of word pairs and human-rated similarity score to each pair. SCWS dataset consists of 2003 word pairs and their sentential contexts. It is worth noting that the scores to each word pairs in WS-353 are given without any contextual information, thus result on SCWS dataset is more reliable.

To ensure a fair comparison with~\newcite{neelakantan2015efficient}, we use the word embeddings released by them. For evaluation metrics, we adopt avgSim on WS-353 and localSim (also referred to maxSimC proposed by \newcite{reisinger2010multi}) on SCWS dataset, respectively. 
In Table~\ref{table: result}, we report the Spearman rank correlation between similarity scores of models and the human judgments in the datasets.
$V$ refers to the 300-dimensional non-parametric multi-sense skip-gram model \cite{neelakantan2015efficient}, and $\tilde{V}_{\rm WordNet}$ refers to the method in \newcite{shi2016real}. 
Our method boosts performance by remarkable $5.6$ points, even better than semi-supervised method. 
PCA and Ex-RPCA achieves almost the same results on two dataset. 
While equipped with the real multi-sense indicator, Ex-RPCA has a wider field of application with good prospects.

\begin{table}[t]
\centering
\begin{tabular}{lrrr}
\hline
\textbf{Vector} & \textbf{WS353} & \textbf{SCWS} & \textbf{${\rm \mathbf{Rank}}(L)$} \\ 
\hline 
$V$ & 68.6 & 59.8 & N/A\\
$\tilde{V}_{\rm WordNet}$ & 69.1 & 62.3 & N/A \\
      $\tilde{V}_{\rm PCA}$ & \textbf{69.2} & \textbf{65.3} & 5 \\
$\tilde{V}_{\rm Ex-RPCA}$ & \textbf{69.2} & \textbf{65.4} & 3 \\
\hline
\end{tabular}
\caption{\label{table: result} Spearman rank correlation ($\rho \times 100$) on Word Sim 353 dataset and SCWS dataset. $V$ refers to the 300-dimensional non-parametric multi-sense skip-gram model \cite{neelakantan2015efficient}, and the results of $\tilde{V}_{\rm WordNet}$ are extracted from \newcite{shi2016real}. ${\rm Rank}(L)$ is the dimensionality of $L$ when reaching the best performance for each type of vectors.}
\vspace{-0.5cm}
\end{table}

\section{Conclusion}
In this paper, we frame the pseudo multi-sense detection into a dimensionality reduction problem. Through proposed Ex-RPCA for principal component analysis, we demonstrate that unsupervised multi-sense word embedding models produce pseudo multi-senses systematically. Moreover, the multi-sense word embeddings can be improved by a simple linear transformation based on Ex-RPCA. Our method boosts performance of the baseline by a large margin. We expect future applications of the proposed method on linguistic analysis for multi-sense word embeddings in the future.


\bibliography{naaclhlt2018}
\bibliographystyle{acl_natbib}

\cleardoublepage
\appendix
{\centering \Large \textbf{Supplemental Materials}}
\section{Proof of the Existence and Uniqueness of Linear Transformation in Section \ref{sec: linear-transformation}}
\label{append: pseudo-elimination}
\subsection{Problem}
Let $W$ be a subspace of $\mathbb{R}^n$, and $\alpha_1,\alpha_2,\cdots,\alpha_k$ be a group of orthogonal basis of $W$. 
We will prove that there exists a unique matrix $T$, which satisfies 
\begin{equation}
T\mathbf{x} = \left\{\begin{aligned} 
\mathbf{0}, & ~~ \mathbf{x} \in W \\
\mathbf{x}, & ~~ \mathbf{x} \in \mathbb{R}^n-W
\end{aligned}
\right.
\end{equation}

\subsection{Proof}
It is easy to extend $\alpha_1, \alpha_2, \cdots, \alpha_k$ to $\alpha_1, \alpha_2, \cdots, \alpha_k, \alpha_{k+1}, \cdots, \alpha_n$, which are a group of orthogonal basis of $\mathbb{R}^n$. 
Let 
\begin{equation}
A = \left(\begin{aligned}
&\alpha_1^T \\
&\alpha_2^T \\
&\cdots \\
&\alpha_n^T
\end{aligned}
\right) = \left(\begin{aligned}
\alpha_{1,1}~~ \alpha_{1,2} ~~ &\cdots ~~ \alpha_{1,n} \\
\alpha_{2,1}~~ \alpha_{2,2} ~~ &\cdots ~~ \alpha_{2,n} \\
&\cdots \\
\alpha_{n,1}~~ \alpha_{n,2} ~~ &\cdots~~ \alpha_{n,n} \\
\end{aligned}
\right)
\end{equation}

\begin{equation}
\left\{
\begin{aligned}
T\alpha_1 & = 0 \\
T\alpha_2 & = 0 \\
& \cdots \\
T\alpha_k & = 0 \\
T\alpha_{k+1} & = \alpha_{k+1} \\
 & \cdots \\
T\alpha_n & = \alpha_n
\end{aligned}
\right. \Leftrightarrow 
\left\{
\begin{aligned}
\alpha_1^T T^T_i & = 0 \\
\alpha_2^T T^T_i & = 0 \\
 & \cdots \\
\alpha_k^T T^T_i & = 0 \\
\alpha_{k+1}^T T^T_i & = \alpha_{k+1, i} \\
& \cdots\\
\alpha_n T^T_i & = \alpha_{n, i} \end{aligned}
\right. 
\forall i, 1\leq i\leq n
\label{eq: matrix-transformation}
\end{equation}
Here, $T_i^T$ denotes the i$^{th}$ column of $T^T$. 
The right part of Eq \eqref{eq: matrix-transformation} can be rewrite to 
\begin{equation}
\centering
A T_i = \overbrace{(\underbrace{0, 0, \cdots, 0}_{k\times 0}, \alpha_{k+1, i}, \cdots, \alpha_{n, i})^T}^{\mathbf{c}_i}
\label{eq: solution}
\end{equation}
Let $\mathbf{c}_i$ denote the right part in Eq \eqref{eq: solution}. 
There exists a unique solution for $T_i=A^{-1}\mathbf{c}_i$ since ${\rm rank}(A) = n$, $\forall i, 1\leq i \leq n$. 
Thus, there exists a unique solution for $T$. \#

\section{Solution for Extended Robust PCA via Convex Optimization}
\label{append: solution}
The augmented Language equation for problem \eqref{eq:convex-formulation} can be written as:
\begin{equation}
\begin{aligned}
&\mathcal{L}(L, E, S, Y) = \sum_{i=1}^{m_0} \sigma_i + \lambda_1 ||E||_F^2 + \lambda_2 ||S||_1 + \\
& \left<Y, M-L-E-S\right> + \frac{\mu}{2}||M-L-E-S||_F^2  
\end{aligned}
\end{equation}
where $Y$ is the Lagrange multiplier, $\left<X,Y\right>$ represents the inner product of two matrices $X$ and $Y$, $m$ is the number of singular values of $L$. 

Before solving the problem, we first define two operators $R_a$ and $D_a$:
\begin{equation}
R_a(X) = {\rm sgn}(X) \max(|X| - a, 0)
\end{equation}
\begin{equation}
\begin{aligned}
D_a(X) & = UR_a(\Sigma)V^T \\
s.t. X & = U\Sigma V^T
\end{aligned}
\end{equation}
where a is a real number, $X$ is a matrix, $U, \Sigma, V$ is the result of singular value decomposition of $X$.

In each iteration $t$ we update $L, E, S$ and $Y$ according to the following equations.
For the convenient of express, all the variables without superscript denote the value at the $t^{th}$ iteration. 
\begin{equation}
\begin{aligned}
&L^{(t+1)} \\ 
&= \argmin_L \sum_{i=1}^m \sigma_i + \frac{\mu}{2} ||M-L-E-S+\frac{1}{\mu}Y||_F^2 \\
&= D_{\frac{1}{\mu}}(M-L-E-S+\frac{1}{\mu}Y)
\end{aligned}
\label{eq: L-update}
\end{equation}
\begin{equation}
\begin{aligned}
&E^{(t+1)} \\
&=\argmin_E \lambda_1 ||E||_F^2 +\frac{\mu}{2} ||M-L-E-S+\frac{1}{\mu}Y||_F^2 \\
&=\frac{\mu}{\mu+2\lambda_1}(M-L-E-S+\frac{1}{\mu}Y)
\end{aligned}
\label{eq: E-update}
\end{equation}
\begin{equation}
\begin{aligned}
&S^{(t+1)} \\
&= \argmin_S \lambda_2 ||S||_1 +\frac{\mu}{2} ||M-E-L-S+\frac{1}{\mu}Y||_F^2 \\
&= R_{\frac{\lambda_2}{\mu}}(M-L-E-S-\frac{1}{\mu}Y)
\end{aligned}
\label{eq: S-update}
\end{equation}
\begin{equation}
\begin{aligned}
& Y^{(t+1)} = Y + \mu(M-L-E-S)
\end{aligned}
\label{eq: Y-update}
\end{equation}

According to \newcite{lin2011linearized}, we design the update policy for $\mu$ as
\begin{equation}
\mu^{(t+1)} = \left\{
\begin{aligned}
& \rho \mu ~~ if \sqrt{\mu}\frac{||E^{(t+1)}+S^{(t+1)}-E-S||_F}{||M||_F}<\epsilon \\
& \mu ~~~~otherwise
\end{aligned}
\right.
\label{eq: mu-update}
\end{equation}

The procedure to solve this problem is summarized in Algorithm \ref{alg: ialm}.

\begin{algorithm}[t]
\begin{algorithmic}[1]  
\REQUIRE{sense-wise difference matrix $M$, weight terms $\lambda_1, \lambda_2$ \\}
\ENSURE{low-rank matrix $L$, sparse noise matrix $S$, Gaussian noise matrix $E$\\}
\STATE{Initialize $\mu^{(0)} by \frac{0.5}{||{\rm sgn}(M)||_2}$}
\STATE{Initialize $\rho$ by 6}
\STATE{Initialize $t$ by 0}
\STATE{Random initialize $L, E, Y, S$}
\WHILE{{\sl not converged}}
 \STATE{Update $L^{(t+1)}$ by Eq \eqref{eq: L-update}}
 \STATE{Update $E^{(t+1)}$ by Eq \eqref{eq: E-update}}
 \STATE{Update $S^{(t+1)}$ by Eq \eqref{eq: S-update}}
 \STATE{Update $Y^{(t+1)}$ by Eq \eqref{eq: Y-update}}
 \STATE{Update $\mu^{(t+1)}$ by Eq \eqref{eq: mu-update}}
 \STATE{Let $M^{(t+1)} = M^{(t)} - S^{(t)}$}
 \STATE{Let $t = t + 1$}
\ENDWHILE
\STATE{\textbf{return} $L^{(t)}, E^{(t)}, S$}
\end{algorithmic}
\caption{Solution for Ex-RPCA via convex optimization.}
\label{alg: ialm}
\end{algorithm}

\newpage

\section{An Overview of Pseudo Multi-Sense}
\begin{figure}[t]
\centering
\includegraphics[width=0.5\textwidth]{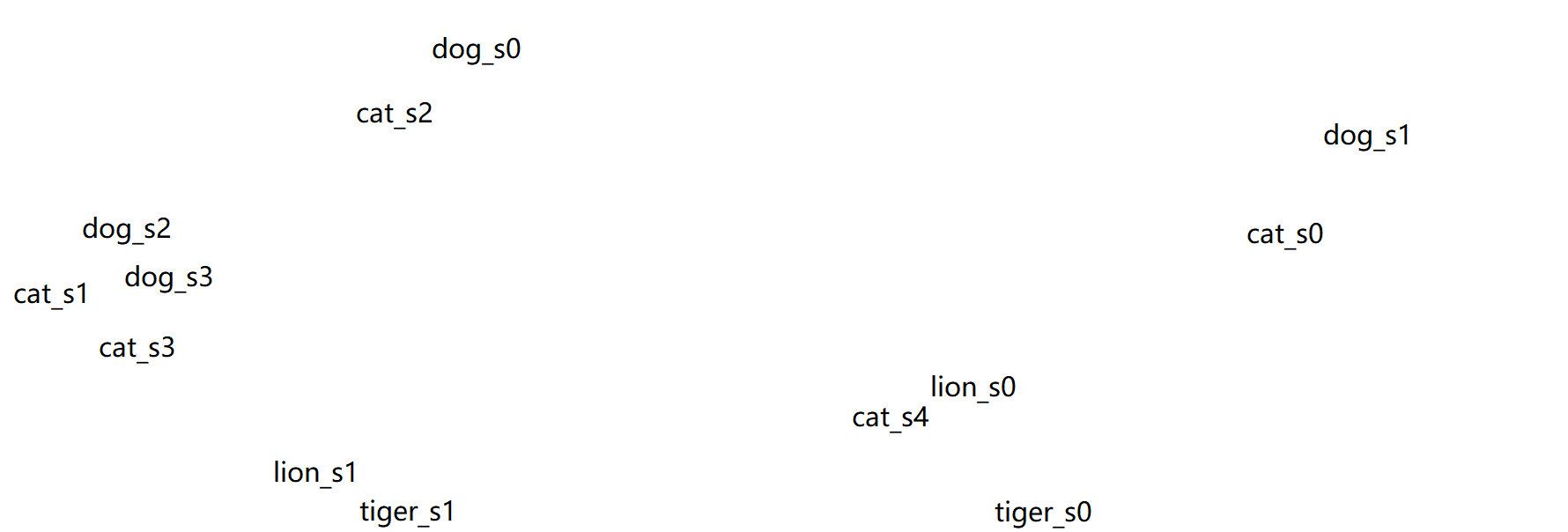}
\begin{tabular}{p{0.47\textwidth}}
\small
Example Sentences \\
\hline
\small
The domestic \textbf{cat\_s0} is a small, typically furry, carnivorous mammal. \\
\small
Tortoiseshell is a \textbf{cat\_s1} coat coloring named for its similarity to tortoiseshell material.\\
\small
Stubbs (a cat's name) was a \textbf{cat\_s2} who was the mayor of Talkeetna, Alaska. \\
\hline
\end{tabular} 
\caption{\label{fig: overview} An overview of \textit{pseudo multi-sense}. 
The figure above shows some word vectors (reduced to 2-D) in multi-sense word embeddings released by \newcite{neelakantan2015efficient}. The example sentences below are selected from Wikipedia by their contextual information of \textit{cat}.
}
\end{figure}

Are pseudo multi-sense generated systematically or accidentally? Figure~\ref{fig: overview} provides a direct evidence. 
In the figure, each \textit{dog} is associated with a \textit{cat}, and each \textit{tiger} is associated with a \textit{lion}.
What is more, the example sentences show that the three senses of \textit{cat} should have the same meaning, \textit{i.e.}, pseudo multi-senses.

\section{More Experimental Results}

\begin{figure}[t]
\centering
\includegraphics[width=0.45\textwidth]{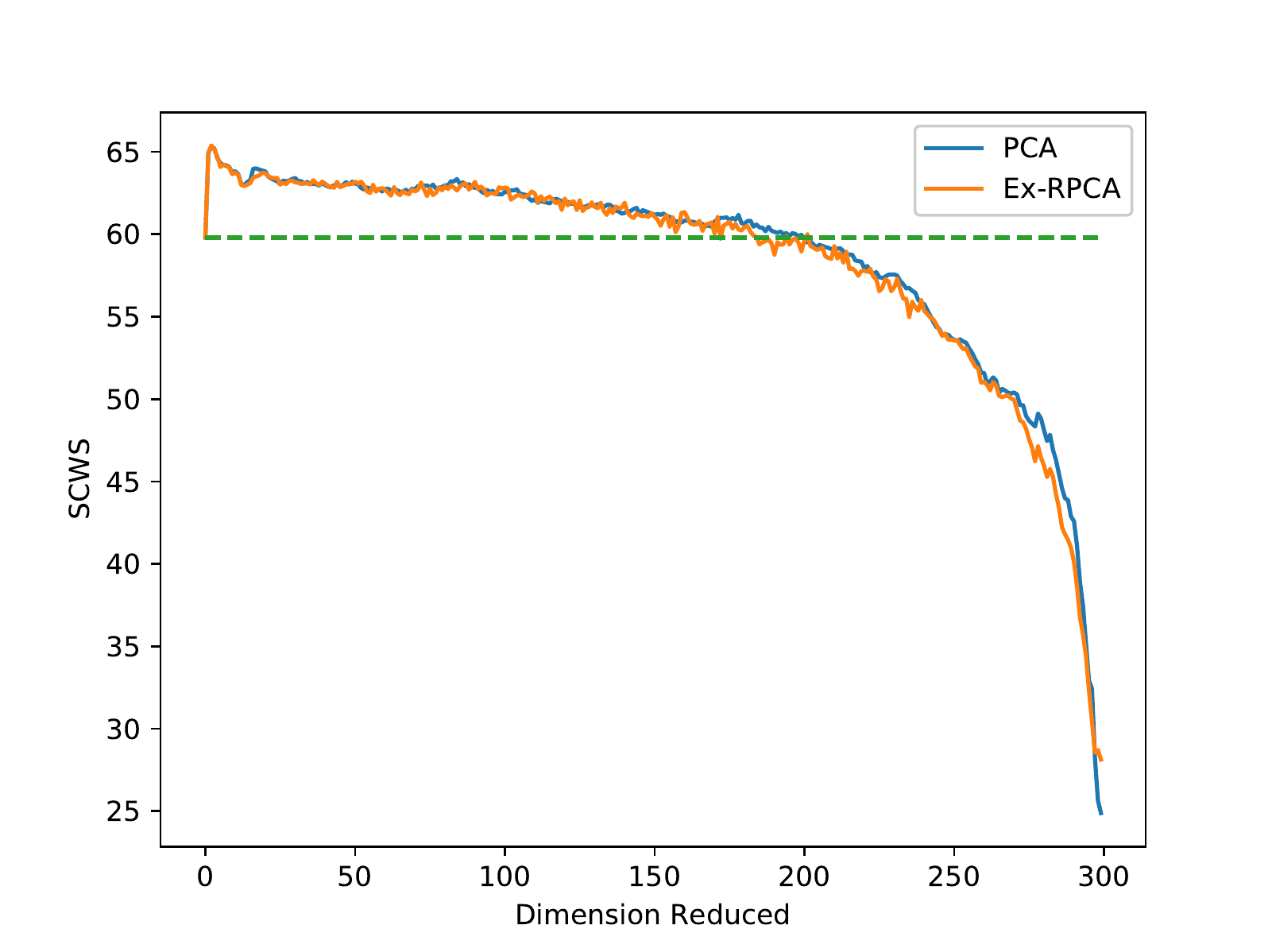} 
\vspace{-0.3cm}
\caption{\label{fig: scws-dim} Dimension reduced-SCWS Spearman correlation curve.}
\end{figure}

Figure~\ref{fig: scws-dim} shows the relation between Spearman rank correlation and the number of dimension reduced in our methods. The performance drops very slowly with the increasing reduced dimensionality at first, but speeds up after reducing two-thirds of the original dimensions. Interestingly, only one-third of its dimensions is needed to keep the original performance on SCWS dataset.

\end{document}